\documentclass[runningheads]{llncs}
\usepackage{amssymb}
\usepackage{graphicx}
\usepackage[T1]{fontenc}

\usepackage{float}
\usepackage{graphicx}
\usepackage{hyperref}
\usepackage{amsmath}
\usepackage{authblk}
\usepackage{multirow}

\begin{document}

\title{Motion-Boundary-Driven Unsupervised Surgical Instrument Segmentation in Low-Quality Optical Flow}

\author{
Yang~Liu\inst{1}\textsuperscript{*} \and
Peiran Wu\inst{2}\textsuperscript{*} \and
Jiayu Huo\inst{1} \and
Gongyu Zhang\inst{1} \and
Zhen Yuan\inst{1} \and
Christos Bergeles\inst{1} \and
Rachel Sparks\inst{1} \and
Prokar Dasgupta\inst{1} \and
Alejandro Granados\inst{1} \and
Sebastien Ourselin\inst{1}
}
%
\authorrunning{Liu et. al.}

\institute{King's College London, London, UK \and University of Bristol, Bristol, UK
}

\maketitle        %

\begingroup
\renewcommand\thefootnote{*}
\footnotetext{Equal contribution.}
\endgroup
\begin{abstract}

Unsupervised video-based surgical instrument segmentation has the potential to accelerate the adoption of robot-assisted procedures by reducing the reliance on manual annotations. However, the generally low quality of optical flow in endoscopic footage poses a great challenge for unsupervised methods that rely heavily on motion cues. To overcome this limitation, we propose a novel approach that pinpoints motion boundaries, regions with abrupt flow changes, while selectively discarding frames with globally low-quality flow and adapting to varying motion patterns. Experiments on the EndoVis2017 VOS and EndoVis2017 Challenge datasets show that our method achieves mean Intersection-over-Union (mIoU) scores of 0.75 and 0.72, respectively, effectively alleviating the constraints imposed by suboptimal optical flow. This enables a more scalable and robust surgical instrument segmentation solution in clinical settings. The code will be publicly released.

\end{abstract}

\section{Introduction}

Instrument segmentation is a key component of robotic-assisted surgeries, providing improved guidance and supporting decision-making. It is also essential for some other AI-driven surgical tasks, such as workflow recognition \cite{liu2023skit,liu2023lovit,shi2021semi}, action identification \cite{psychogyios2023sar}, and tracking \cite{bouget2017vision}.

Although deep learning has made significant progress in fully- \cite{9975835,ni2022surginet,ni2019rasnet,shvets2018automatic,jin2019incorporating} and semi-supervised \cite{wei2023segmatch,wu2024surgivisor} segmentation, these approaches rely heavily on manual annotations, which increases the workload and limits their application. In contrast, we focus on unsupervised segmentation to enhance surgery understanding without any additional manual effort. Early unsupervised approaches in surgical tasks mostly focuesd on workflow analysis \cite{bodenstedt2017unsupervised} and motion prediction \cite{dipietro2018unsupervised}, while instrument segmentation has received less attention. Liu et. al. \cite{liu2020unsupervised} used unsupervised techniques and basic handcrafted cues for segmentation. However, their approach's reliance on specific signals limits its adaptability across different surgeries.

Motion plays an essential role in visual processing and is widely utilised in unsupervised models like RAFT \cite{teed2020raft} to effectively capture movement in videos. Sestini et. al. \cite{sestini2023fun} introduced FUN-SIS, which integrates realistic instrument segmentation masks (referred to as \textit{shape-priors}) from various datasets with optical flow information. Despite its novelty, FUN-SIS's dependency on the \textit{shape-priors} and its non-end-to-end nature limited its applicability. Recently, Lian et. al. \cite{Lian_2023_CVPR} brought forward RCF, a fully unsupervised model for video object segmentation (VOS), demonstrating the critical role of motion. The success of RCF highlights the importance of optical flow quality in unsupervised learning, establishing a new benchmark for instrument segmentation without manual annotations.

\begin{figure}[t]
\centering
\includegraphics[width=\textwidth]{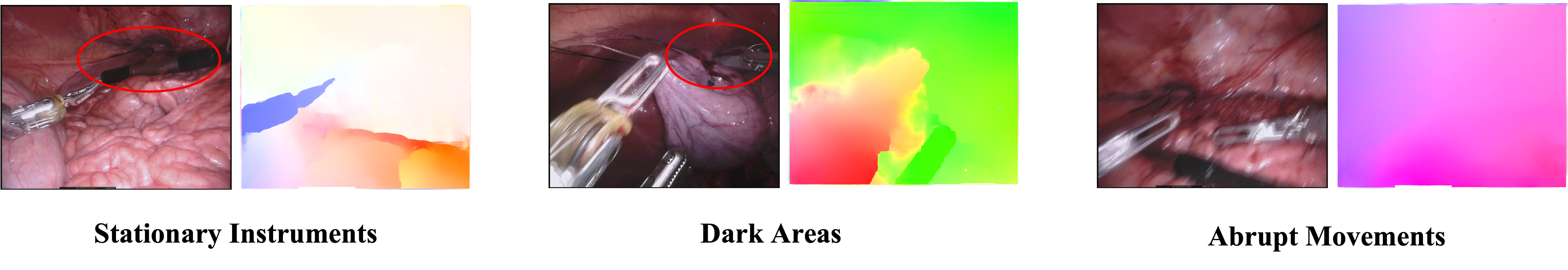}
\caption{Example of some low-quality optical flow frames, including stationary instruments, dark areas and abrupt movements, which greatly limit the model performance.} \label{fig1}
\end{figure}

Unlike the high-quality optical flow found in natural videos, surgical footage often suffers from low flow quality due to the factors such as dark areas, abrupt movements, and stationary instruments, as shown in Fig.~\ref{fig1}. Following the approach of the RCF model~\cite{Lian_2023_CVPR}, we aim to reduce the impact of low-quality optical flow on our model. Liao et al.~\cite{liao2023mobile} highlighted the importance of accurately segmenting boundaries, a task complicated by the blurred boundaries in surgical optical flow. Inspired by Super-BPD~\cite{Wan_2020_CVPR}, we employ angular measurements to detect these unclear boundaries in the flow, ensuring the model focuses on the most reliable region. Additionally, the frequent presence of low-quality optical flow in surgery videos presents a notable challenge. To address this, we drop the most difficult samples in each batch. Furthermore, the subtle movement of certain instruments, often not captured by optical flow, can reduce the model sensitivity. By varying frame rates beyond the standard, we enhance the detection of these less visible movements.

\section{Method}

We target the challenge of unsupervised surgical instrument segmentation within a video stream of $T$ frames, denoted as $X_T = \{\boldsymbol{x}_i|i \in \{1,..., T\}\}$, where $\boldsymbol{x}_i$ represents the image of the $i$-th frame. 
The objective is to create a mapping function $f$ such that $f(x_i)$ produces a mask that extracts the area of the surgical instrument. It is important to note that instrument mask annotations are not employed during the training stage.

\begin{figure}[t]
\centering
\includegraphics[width=\textwidth]{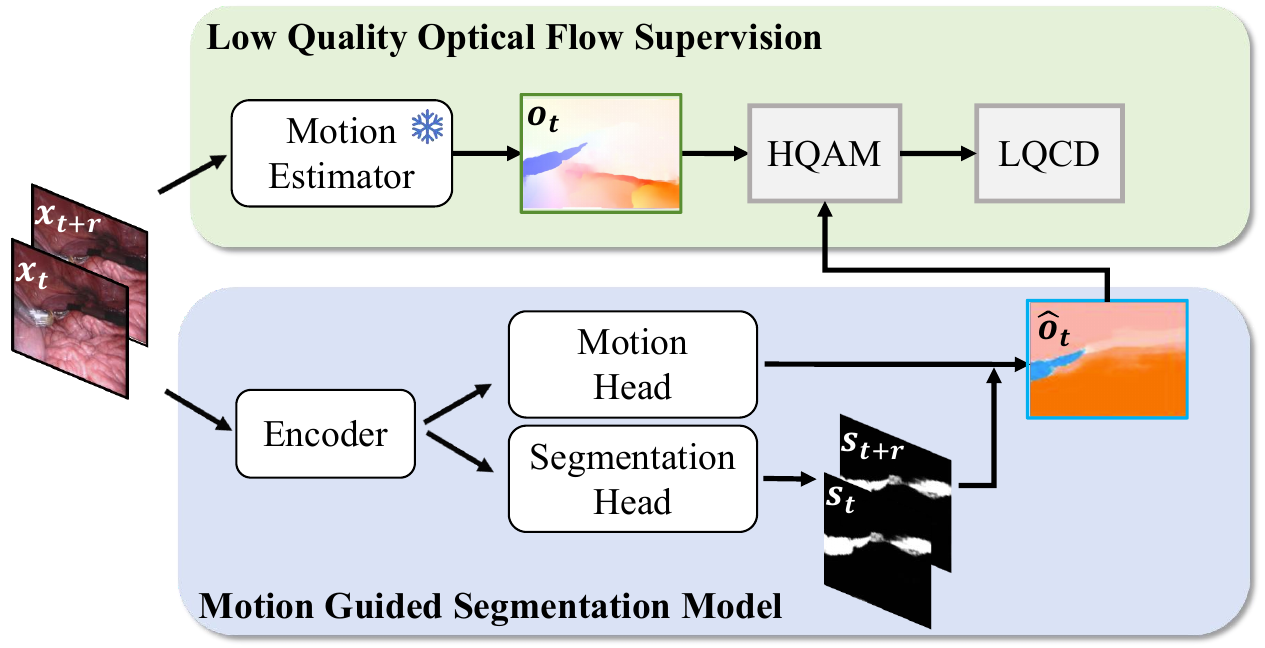}

\caption{Overview of our proposed unsupervised instrument segmentation framework. Two frames, separated by a random interval \(r\), are fed into both a motion-guided segmentation model (e.g.~RCF~\cite{Lian_2023_CVPR}) and a pre-trained Motion Estimator ( e.g.~RAFT~\cite{teed2020raft}) that generates pseudo flow maps $o_t$. The proposed HQAM and LQCD modules refine these pseudo flow maps, yielding a robust supervision.} \label{fig2}
\end{figure}

We adopt the first stage of the state-of-the-art (SOTA) RCF model~\cite{Lian_2023_CVPR} as our backbone, thanks to its fully end-to-end design and independence from \emph{shape-priors}. However, the optical flow in surgical videos often suffers from dim lighting, abrupt instrument or camera movements, and partially stationary objects (Fig.~\ref{fig1}), creating significant challenges for unsupervised segmentation. To address these issues, we introduce a \textbf{High-Quality Area Matching (HQAM)} block that emphasizes reliable flow regions, a \textbf{Low-Quality Case Dropping (LQCD)} mechanism to discard frames with severely compromised flow, and a variable frame-rate scheme to better capture subtle instrument motions. The overall architecture of our method is illustrated in Fig.~\ref{fig2}.

\begin{figure}[t]
\centering
\includegraphics[width=\textwidth]{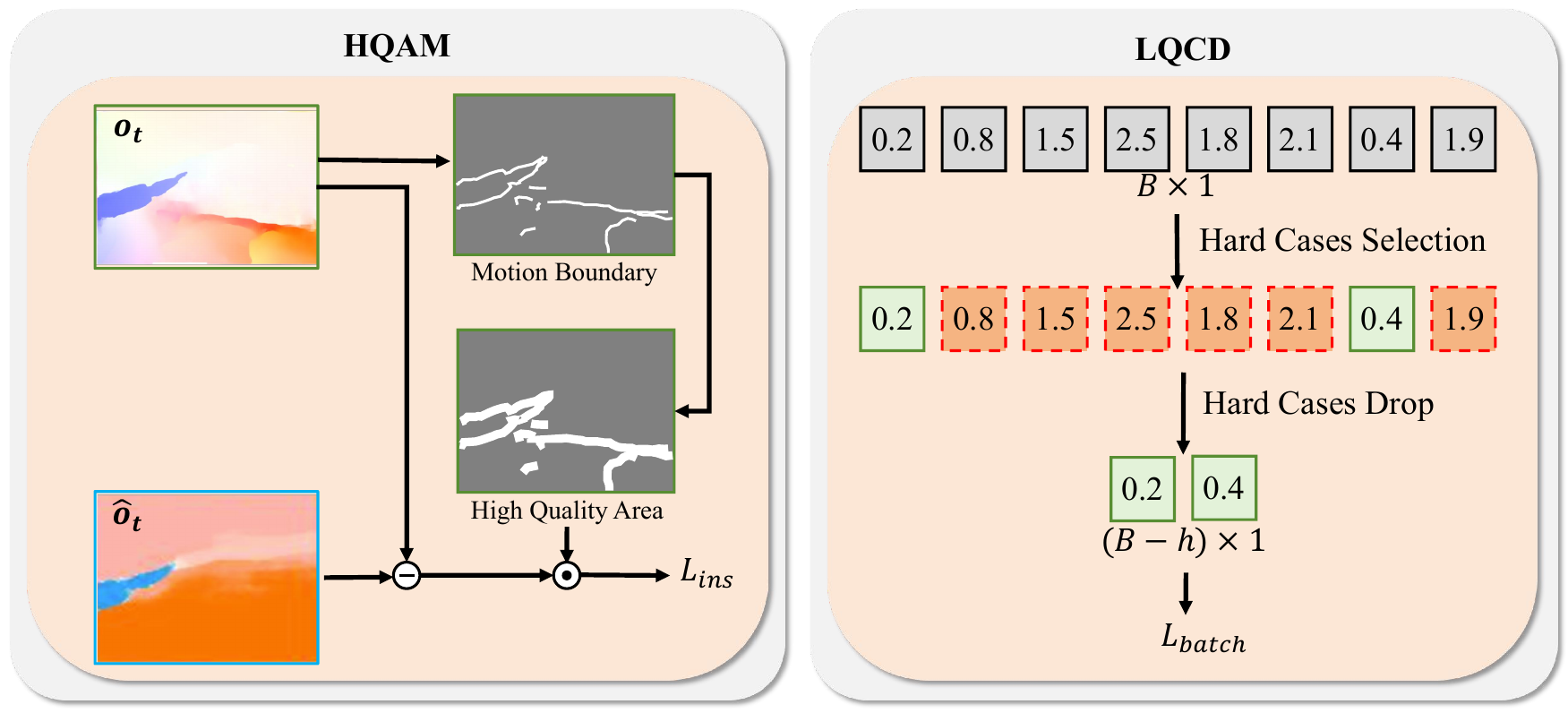}

\caption{Illustration of our HQAM and LQCD modules. HQAM derives a boundary-based mask from pseudo optical flows $o_t$, isolating reliable \emph{high-quality} regions to guide segmentation. Meanwhile, LQCD ranks each frame in a batch by its per-frame loss and discards the top \(h\) ``hard cases'', removing globally low-quality motion signals.}\label{fig_module}
\end{figure}

\subsection{High-Quality Area Matching}

Accurate motion cues derived from optical flow are essential for unsupervised instrument segmentation. However, in surgical videos, these cues often fail to capture the interior of instruments, particularly in low-light or partially stationary scenarios (Fig.~\ref{fig1}), resulting in missed motion signals. If such incomplete cues are used for training, they risk propagating errors. In contrast, regions closed to the flow changes abruptly (\emph{motion boundaries}) typically provide more reliable signals. Building on this insight, we propose a novel High-Quality Area Matching (HQAM) module, as shown in Fig.~\ref{fig_module}, that selectively emphasizes these high-contrast motion areas. By directing supervision toward the most trustworthy regions, HQAM mitigates the impact of missed interior motion and significantly improves segmentation performance.

Initially, we transform the optical flow representation $o_i \in \mathbb{R}^{H \times W \times 2}$, encoding motion vectors in two dimensions (horizontal and vertical movements), into a directional format $\theta_i \in \mathbb{R}^{H \times W}$, where $H$ and $W$ represent the height and width of the image frame, respectively. This conversion calculates the angle $\theta_i^p$ of the optical flow at each pixel position $p = (j,k)$. Specifically, the direction angle $\theta_i^p$ at position $p$ is obtained through the formula:
\[
\theta_i^p = \arctan2(o_i^{j,k,x}, o_i^{j,k,y}),
\]
where $o_i^{j,k,x}$ and $o_i^{j,k,y}$ are the horizontal and vertical components of the optical flow vector at the pixel $p$, respectively. Next, we define the directional difference at each pixel $p$ as follows:
\[
\delta_i^p = \max_{n^p \in \mathcal{N}_1^p} |\theta_i^p - \theta_i^{n^p}|,
\]
where $\mathcal{N}_1^p$ signifies the 4-neighbourhood of pixel $p$, including the adjacent pixels in both the vertical and horizontal orientations. To demarcate boundary regions, we introduce a difference threshold $\alpha = \frac{\pi}{12}$. This leads to the creation of the boundary mask ${M}_i$, where a value of 1 is assigned to pixels that meet the condition $\delta_i^p > \alpha$, indicating the presence of a boundary, while a value of 0 denotes areas without boundaries. To enhance the supervision diversity, a simple dilation technique is employed to broaden the areas under supervision. Specifically, for every boundary pixel satisfying $\delta_i^p > \alpha$, we set $M_i(\mathcal{N}_d^p)$ to 1, where $d$ represents the dilation distance. Consequently, we define the instance-level optical flow loss by averaging the per-pixel loss over the entire image domain \(\Omega\):
\[
\mathcal{L}_{\mathrm{ins}} \;=\; \frac{\sum_{p \in \Omega} M_i(p)\,\bigl\lVert o_i(p) \;-\; \hat{o}_i(p)\bigr\rVert^2}{\sum_{p \in \Omega} M_i(p)} .
\]

\subsection{Low-Quality Cases Drop}

While HQAM addresses local flow inaccuracies, we observe that low-quality optical flow often spans the entire frame rather than being confined to specific regions. In other words, frames with degraded local flow typically exhibit inferior quality overall. To tackle this issue, we propose a frame-level drop mechanism Low-Quality Cases Drop (LQCD), as shown in the right of Fig.~\ref{fig_module}, that discards globally problematic frames. Specifically, for a batch of \( B \) training images, we identify and remove the top \(h\) frames with the highest losses (the “hard cases”), leaving us with a subset \(\mathcal{S}_\mathrm{remain}\) of \( B - h \) frames. Our final batch-level loss is then computed as:

\[
\mathcal{L}_\mathrm{batch} 
= \frac{1}{\lvert \mathcal{S}_\mathrm{remain}\rvert}
  \sum_{x_i \,\in\, \mathcal{S}_\mathrm{remain}}
  \mathcal{L}_\mathrm{ins}(x_i).
\]
By removing globally low-quality flow cases, we strengthen the reliability of frame-level supervision and reduce error propagation.

\subsection{Variable Frame Rates Training Input}

Despite establishing reliable supervision mechanisms, instruments that frequently remain stationary pose a challenge to effective training due to the scarcity of meaningful optical flow supervision samples in these regions. To overcome this limitation, we introduce a strategy of feeding training images with variable frame rates, $(x_i, x_{i+r})$, where the interval between adjacent frames, $r$, is a random number within the range of $1$ to $3$, instead of being fixed at $r = 1$. This variability ensures that the optical flow can capture the motion of instruments that typically exhibit little or no movement, thereby facilitating more effective training.

\section{Experiments}
\subsection{Datasets and Evaluation Metrics}
We implemented our experiments on datasets from the MICCAI EndoVis 2017 Robotic Instrument Segmentation Challenge. This dataset features 8 sets, each with 225 frames of stereo camera footage recorded by the da Vinci Xi surgical robot during various pig surgery procedures. Each frame has been meticulously annotated by experts, with identifed distinct parts of robotic surgical instruments. These instruments have been further categorised into rigid axes, articulated wrists, grommets, and a miscellaneous category for additional instruments like laparoscopic tools or drop-in ultrasound probes. We explored two variations of the EndoVis 2017 dataset for our study: EndoVis 2017 VOS and EndoVis 2017 Challenge. The VOS version includes laparoscopic tools and ultrasound probes as per Sestini et al. \cite{sestini2023fun}, while the Challenge version adheres to the original competition guidelines by excluding the `other label' category. In the evaluation, we followed the same conventions and the same data partitioning and performed a 4-fold cross-validation on both the VOS and the challenge versions of the dataset~\cite{jin2019incorporating}. Our main evaluation metric is the mean Intersection-over-Union (mIoU), consistent with the benchmarks set in the MICCAI EndoVis 2017 Challenge and subsequent research in the field.

\subsection{Implementation Details}
We chose the RCF framework~\cite{Lian_2023_CVPR} as the backbone for our model, leveraging a ResNet50~\cite{he2016deep} for feature extraction, which feeded into both a segmentation and a residual prediction head. The optical flow was derived using the RAFT model, pretrained on synthetic datasets - FlyingChairs~\cite{dosovitskiy2015flownet} and FlyingThings~\cite{mayer2016large}, without any human-annotated data. Our implementation was based on PyTorch and runs on a single NVIDIA A100-PCIE-40GB GPU. The batch size was set to 8, with the $h = 6$ hard cases to ensure frame-level supervision quality. For boundary dilation, a kernel size of $d = 7$ was used.

\subsection{Comparison with State-of-the-art Methods}

\begin{table}[t]
\centering
\renewcommand{\arraystretch}{1.2}
\caption{Comparison with SOTA supervised methods and unsupervised methods. Mean IoU (\%) is reported. And the results of prior works are quoted.}
\resizebox{\linewidth}{!}{
\begin{tabular}{cccccc}
\hline
Annotation & \textit{Shape-priors} & Stage & Method & EndoVis 2017VOS & EndoVis 2017Challenge \\
\hline
100\% & No & 1 & TernausNet~\cite{shvets2018automatic} & 89.06 & 83.60 \\
100\% & No & 1 & MF-TAPNet~\cite{jin2019incorporating} & 89.61 & 87.56 \\
\hline
0\% & Yes & 1 & FUN-SIS(Stage1) & 40.08 & 37.03 \\
0\% & Yes & 2 & FUN-SIS(Stage2) & 74.78 & 68.31 \\
0\% & Yes & 3 & FUN-SIS(Stage3)\cite{sestini2023fun} & 83.77 & 76.25 \\
\hline
0\% & No & 1 & AGSD~\cite{liu2020unsupervised} & 71.47  & 67.85 \\
0\% & No & 1 & RCF(Stage1) & 46.09 & 46.17 \\
0\% & No & 2 & RCF(Stage2)\cite{Lian_2023_CVPR} & 49.18 &  49.19\\
0\% & No & 1 & RCF(Stage1) + \textbf{Ours} & \textbf{75.07} & \textbf{72.07} \\
\hline
\label{table1}
\end{tabular}}
\end{table}

\begin{figure}[t]
\centering
\includegraphics[width=0.8\textwidth]{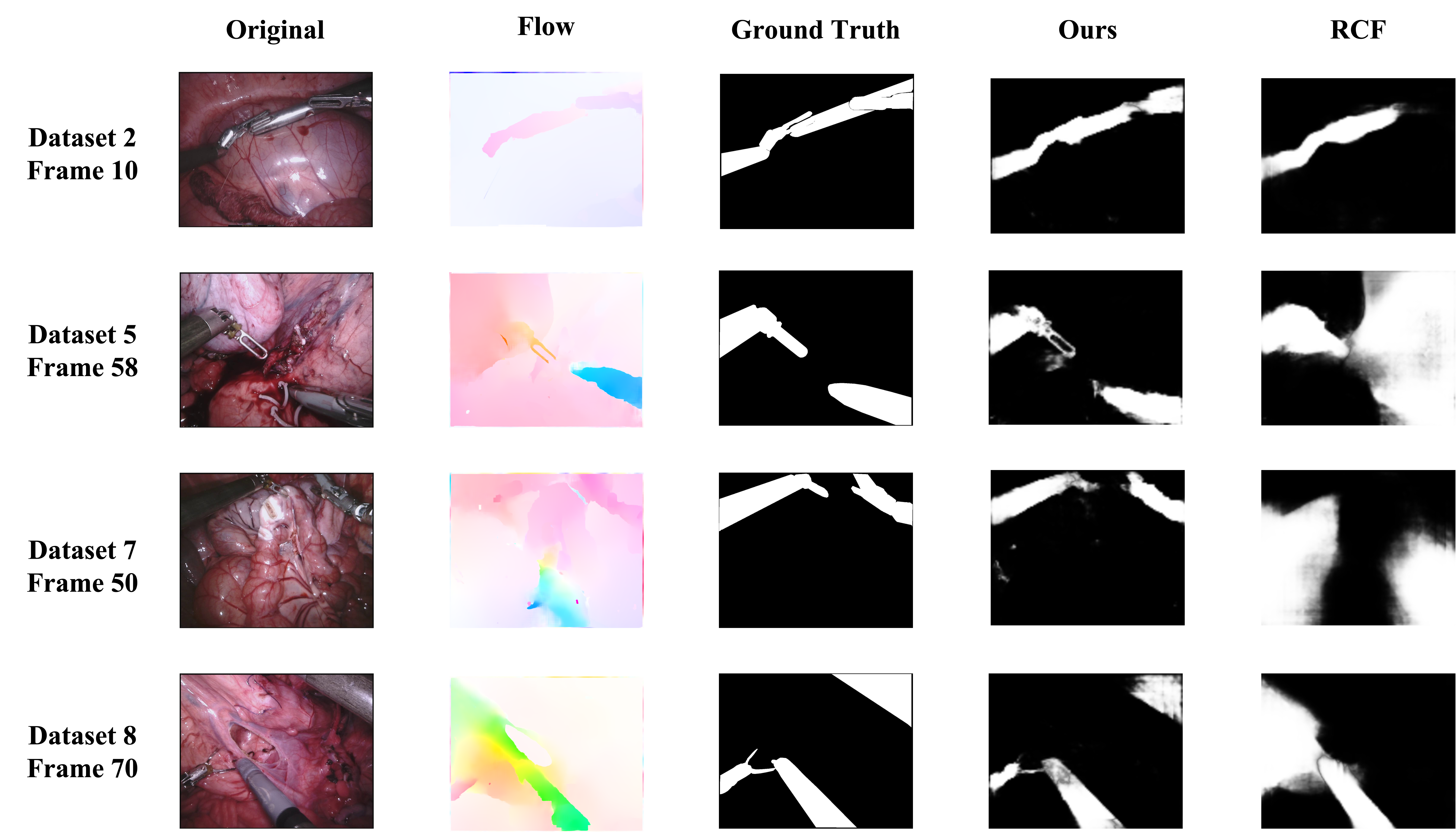}
\caption{Qualitative comparisons with the baseline model RCF, showing (a) optical flow pseudo-labels obtained by RAFT prediction (b) Ground Truth from EndoVis 2017, offering (c) Prediction masks of our method (d) Prediction masks of RCF.} \label{fig3}
\end{figure}

Our findings on the EndoVis 2017 VOS and EndoVis 2017 Challenge datasets are presented in Table~\ref{table1}. We compared our method with the AGSD~\cite{liu2020unsupervised} and FUN-SIS~\cite{sestini2023fun} methods, which were both tailored to surgical video segmentation, and RCF~\cite{Lian_2023_CVPR}, that was originally designed to segment natural images.
Our method advanced the RCF model, delivering an end-to-end solution without relying on \textit{shape-priors} for annotations. The enhancements we had implemented lead to substantial performance enhancements, showing a 28.98 percentage points (pp) and 25.93 pp increase over the original RCF model on the EndoVis 2017 VOS and Challenge datasets, respectively. This demonstrated our approach's capability to refine the use of low-quality optical flow.
Moreover, when measured against AGSD, our fully unsupervised method presented gains of 3.6 pp and 4.22 pp for the respective datasets, marking the first instance of a purely unsupervised model using optical flow outperforming those dependent on crafted pseudo-labels.
Our model also showed an edge over the FUN-SIS's second stage, surpassing it by 0.29 pp and 3.76 pp on the VOS and Challenge versions, respectively.

Fig.~\ref{fig3} showcases comparative visual examples between our method and the baseline RCF model. Although RCF struggles with the typically low optical flow in surgical videos, our model excels, showing impressive results even in areas with challenging flow conditions. Particularly in detailed areas, such as the instrument head at case frame 58, our approach not only matches but also surpasses the ground truth.

\subsection{Ablation Study}
To ensure fairness and generalisability, all our experiments employ a 4-fold cross-validation on the EndoVis 2017 VOS dataset. We conducted extensive ablation studies to evaluate the impact of our proposed three key components: HQAM, LQCD and variable frame rates training input (Variable).

\noindent \textbf{Different Components of the Method.} Table~\ref{table2} presents the ablation study results of our proposed approach. Starting with the baseline model (first row), we observe a modest performance of 46.09\% mIoU. Introducing LQCD alone raises the mIoU to 47.15\%, confirming that discarding frames with severely degraded optical flow helps reduce noisy supervision. When we further integrate HQAM, the performance jumps significantly to 74.47\% mIoU. This demonstrates the importance of leveraging more reliable boundary cues for instrument segmentation in unsupervised settings. Finally, adding the variable frame-rate strategy produces a slight yet consistent improvement, achieving our best result of 75.07\% mIoU, which indicates that adapting to different temporal intervals provides complementary gains. Overall, these findings highlight the effectiveness of each component and underscore their synergistic contributions to improving unsupervised instrument segmentation.

\noindent \textbf{Parameter Analysis of HQAM.} We investigate the impact of two key HQAM hyperparameters: the angle threshold \(\alpha \in \bigl\{\tfrac{\pi}{12}, \tfrac{\pi}{6}, \tfrac{\pi}{3}\bigr\}\) and the dilation kernel size \(d \in \{1, 3, 7\}\). In these experiments, HQAM is the only enhancement used (i.e., both LQCD and variable frame rates are disabled). As shown in Table~\ref{table3}, the best performance (70.66\%) is achieved at \(\alpha = \tfrac{\pi}{3}\) and \(d=7\), a substantial improvement over the 46.09\% baseline without HQAM. Although performance does vary across hyperparameters, an expected outcome in an \textit{unsupervised} setting, every tested configuration with HQAM significantly surpasses the baseline, highlighting the method’s overall robustness and effectiveness. In our default setting, we adopt more conservative parameters \((\alpha = \tfrac{\pi}{12},\, d=7)\) for broader generalization in diverse surgical scenarios. Nevertheless, the consistent gains under different hyperparameter choices underscore HQAM’s adaptability and confirm its value in unsupervised instrument segmentation.
\begin{table}[t]
\centering
\renewcommand{\arraystretch}{1.2}
\begin{minipage}[t]{0.55\textwidth}
\centering
\caption{The effects of different components in method.}
\resizebox{0.9\linewidth}{!}{
\begin{tabular}{ccc|c}
\hline
\multicolumn{3}{c|}{Function} & Metric \\
LQCD & HQAM & Variable & mIoU(\%)\\
\hline
 & & & 46.09\\
$\checkmark$ & & & 47.15\\
$\checkmark$& $\checkmark$& & 74.47\\
$\checkmark$ & $\checkmark$ & $\checkmark$ & 75.07\\
\hline
\label{table2}
\end{tabular}
}
\end{minipage}
\begin{minipage}[t]{0.4\textwidth}
\centering
\caption{Different parameters of HQAM.}
\resizebox{0.6\linewidth}{!}{
\begin{tabular}{cc|c}
\hline
\multicolumn{2}{c|}{Parameters} & Metric \\
$\alpha$ & $d$ & mIoU(\%)\\
\hline
\multirow{3}{*}{$\frac{\pi}{12}$} & 7 & 68.34\\
 & 3 & 66.04\\
 & 1 & 67.81\\
\hline
\multirow{3}{*}{$\frac{\pi}{6}$} & 7 & 68.74\\
& 3 & 66.03\\
 & 1 & 66.21\\
\hline
\multirow{3}{*}{$\frac{\pi}{3}$} & 7 & \textbf{70.66}\\
 & 3 & 67.03\\
& 1 & 64.74\\
\hline
\label{table3}
\end{tabular}}
\end{minipage}

\end{table}

\section{Conclusion}

In this study, we presented an unsupervised surgical instrument segmentation method that integrates High-Quality Area Matching (HQAM), Low-Quality Cases Dropping (LQCD), and variable frame rates to address the challenges posed by low-quality optical flow. Evaluated on the MICCAI EndoVis 2017 VOS and Challenge datasets, our approach achieves mIoU scores of 0.75 and 0.72, respectively, effectively reducing the need for manual annotations in clinical environments. Notably, our method adopts a plug-and-play design, allowing extension to additional motion-driven tasks such as unsupervised depth estimation. While our experiments indicate certain hyperparameter sensitivities, common in unsupervised settings, this limitation points to a promising area for further research to enhance the robustness and generalizability of our framework.

\bibliographystyle{splncs04}
\bibliography{mybibliography}

\end{document}